# Enhanced Gradient for Differentiable Architecture Search

Haichao Zhang, Kuangrong Hao, *Member, IEEE*, Lei Gao, *Member, IEEE*, Xuesong Tang, and Bing Wei, *Student Member, IEEE*

*Abstract*—In recent years, neural architecture search (NAS) methods have been proposed for the automatic generation of task-oriented network architecture in image classification. However, the architectures obtained by existing NAS approaches are optimized only for classification performance and do not adapt to devices with limited computational resources. To address this challenge, we propose a neural network architecture search algorithm aiming to simultaneously improve network performance (e.g., classification accuracy) and reduce network complexity. The proposed framework automatically builds the network architecture at two stages: block-level search and network-level search. At the stage of block-level search, a relaxation method based on the gradient is proposed, using an enhanced gradient to design high-performance and low-complexity blocks. At the stage of network-level search, we apply an evolutionary multi-objective algorithm to complete the automatic design from blocks to the target network. The experiment results demonstrate that our method outperforms all evaluated hand-crafted networks in image classification, with an error rate of $3.19\%$ on CIFAR10 and an error rate of $19.72\%$ on CIFAR100, both at network parameter size less than one megabit. Moreover, compared with other neural architecture search methods, our method offers a tremendous reduction in designed network architecture parameters.

## I. INTRODUCTION

Over the past decades, convolution neural network (CNN) has shown remarkable success in almost all aspects of computer vision. It is first successfully applied in image classification, where many convolution neural network architectures are developed, including AlexNet [1], GoogLeNet [2], ResNet [3], DenseNet [4], etc. Meanwhile, architectures such as ShuffleNet [5] and MobileNet [6] are designed to enable the actual deployment of a high-performance model on resource-constrained devices. In most cases, CNN needs to be designed based on a specific task, which leads to endless hyperparameters tuning. To alleviate the tedious work, the neural architecture search (NAS) [7-9] is proposed and it is proven as a promising approach to pose the process of designing neural network architecture as an optimization problem [10, 11]. This automatic and efficient approach for designing neural network architecture has a significant impact on the implementation of neural networks.

So far, there are mainly three pathways for NAS. The first popular way to search network architecture is the evolutionary algorithm [12-14]. This method generates new individuals through evolutionary operations (e.g., crossover and mutation) and obtains the best-performing individuals by searching in the population. The reinforcement learning method [15] is the second most popular one, which applies a recurrent network as the controller to generate the network architecture and optimize the controller through reinforcement learning [16, 17]. Although the performance of the above two methods is impressive, the search process is significantly resource-intensive for a large dataset like ImageNet [18]. Therefore, the method based on the gradient search network architecture has attracted wide attention because it greatly accelerates the search process and achieves impressive performance [19]. The third method, DARTS (Differentiable ARchiTecture Search), [20] is a recently proposed approach with outstanding performance. By relaxing the discrete search space to be continuous, DARTS formulates the process of searching optimal neural network architecture as a bilevel optimization method. It obtains the optimized architecture with respect to its performance on the validation set by gradient descent. Although DARTS has achieved outstanding performance with great search acceleration, it still has two deficiencies. DARTS uses a proxy network composed of fewer blocks to search the structure of normal blocks and reduction blocks [20, 21], and then the obtained block structure is applied/shared in all blocks of the target network. In this case, the complexity of the network is not considered in the process of searching for block structure in the proxy network, which leads to the redundancy of block structure complexity. Furthermore, the shared blocks are not optimally designed in the target network, which causes the architecture redundancy of the target network.

In this work, we adopt a new viewpoint to address the aforementioned deficiencies. We propose a multi-objective NAS algorithm—enhanced gradient for differentiable architecture search (EG-DARTS). The proposed approach takes into account both network performance (classification accuracy in this work) and network complexity when calculating the gradient during searching the architecture. Furthermore, to optimize the architecture of the shared block in the target network, we design a multi-objective evolutionary optimization method based on non-dominated sorting to construct the target

This work was supported in part by the Fundamental Research Funds for the Central Universities (2232021A-10), National Natural Science Foundation of China (nos. 61806051, 61903078), and Natural Science Foundation of Shanghai (20ZR1400400).

Haichao Zhang, Kuangrong Hao, Xuesong Tang, Bing Wei are with the Engineering Research Center of Digitized Textile and Apparel Technology, Ministry of Education, and also with the College of Information Science and Technology, Donghua University, Shanghai 201620, China.
L. Gao is now with CSIRO, Waite Campus, Urrbrae, SA 5064, Australia.

network. Our algorithm fully considers the adaptation of the blocks searched in the proxy network to the target network. The designed algorithm is an end-to-end approach to construct the neural networks, which fully considers the high performance of the network while consuming limited computing resources. This algorithm helps CNN to achieve high-performance actual deployment on resource-constrained devices. The key contributions made in this paper are summarized below:

1) Based on the continuously relaxed search space [21, 22], we establish a differentiable architecture search method for dual objectives (high classification accuracy and low network complexity) to construct low-complexity, high-performance CNN blocks.

2) We introduce an evolutionary optimization method based on non-dominated ranking to construct low-complexity, high-performance target networks on CNN shared blocks. The proposed method reduces the gap between the proxy network and the target network and helps designers to build target networks for feasible deployment in devices with limited computational resources.

3) The proposed method provides excellent network search performance and is tested on datasets CIFAR10 and CIFAR100. Additionally, we analyze the relationship among network parameters, floating-point operations (FLOPS), and interface time. The results show that the inference time is directly related to the network depth.

The rest of the paper is organized as follows: Section II summarizes relevant works in the literature. In Section III, we provide a detailed description of the main components of our approach. We describe the experimental setup to validate our approach along with a discussion of the results in Section IV. Finally, we conclude with a summary of our findings and present possible future directions in Section V.

## II. LITERATURE REVIEW

Since the 20th century, research has been devoted to the automation of neural network design. Genetic algorithm [23] or other evolutionary algorithms [12-14] have been applied to search for high-performance architectures, such as the works [14, 24, 25] that evolved the topology and hyperparameters of neural networks. In recent years, there has been an increasing interest in searching neural network architecture, and the methods based on evolutionary computation have continuously shown satisfying performance in doing so [26-28]. Similarly, reinforcement learning-based NAS has also demonstrated excellent performance, such as the pioneering work that adopted the RNN network as the controller to sequentially design the architecture [16]. The gradient-based NAS [20, 21] method utilizes the gradient of the network architecture parameters to optimize the architecture. This method greatly accelerates the search process to obtain a network architecture with favorable performance. The above streams are the main search strategies frequently used in NAS. Next, we review the studies of neural network architecture search for image classification and present some operations in NAS that are utilized as the foundation of our proposed approach.

### A. NAS based on blocks

Deep neural networks have complex discrete architecture parameters and continuous hyperparameters, so optimizing the architecture of deep neural networks is an arduous task. Many studies [29] have simplified the process of NAS into searching the shared structural blocks of networks. Originally, Liu et al. [30] proposed NASNet, which searches for architectural building blocks on a smaller dataset and then transfers the blocks to a larger dataset. The portability of the blocks proposed by NASNet provides new ideas for NAS. Sun et al. [9] used the genetic algorithm to automatically evolve neural network architectures based on the blocks.

The block-based optimization has also been applied to reinforcement learning-based NAS. Zhong et al. [17] proposed BlockQNN, which provides the block-wise network to generate blocks that are stacked to construct the whole target network. BlockQNN utilizes the Q-learning paradigm with the epsilon-greedy exploration approach to build the shared blocks and accelerate the NAS process. Additionally, limiting the search space [30], early-stop strategy [31], progressive searching [32], and weight sharing in architecture were proposed to speed up the search process. Unfortunately, the NAS methods [33] based on reinforcement learning are still resource hungry and suffer from limited search space because of the fixed-length coding of network structure.

Most gradient-based NAS algorithms optimize the neural network architecture through the block-based approach. Liu et al. [20] proposed the DARTS algorithm, which optimizes the target network by optimizing the blocks in the proxy network. As an efficient framework of NAS, DARTS has been widely used in PolSAR image classification [34]. However, searching in the proxy network is low-level due to the disparity between the proxy and the target network, and the shared blocks are not optimally designed in the target network. Direct Sparse Optimization NAS (DSO-NAS) [35] regarded the optimization of the neural network architecture as the view of model pruning. DOS-NAS utilizes a scaling factor to scale the information flow between operations to optimize an initial fully connected block, and then sparse regularization is applied to pure useless connections in the blocks. However, the optimization of network architecture complexity is not involved in DSO-NAS.

### B. Multi-objective for NAS

Multi-objective optimization is concerned with simultaneous optimization where trade-offs exist between two or more conflicting objectives [22, 36, 37]. NEMO, proposed by Kim et al. [38], is one of the earliest multi-objective evolutionary methods for evolving CNN architecture. NEMO applies NSGA-II [39] to maximize network classification accuracy and minimize the inference time. In the NEMO method, the number of output channels from each layer is searched within a restricted space of seven different structures. Based on this work [32], Dong et al. [40] proposed DPP-Net, which takes a compact architecture search space driven by mobile CNN blocks and further improves search efficiency by adopting the resource consumption. LEMONADE reduces computational requirements through a proxy network and allows the newly generated architecture to share parameters with its forerunners, avoiding the consumption of computing resources caused by training the new network from scratch. However, LEMONADE model on the CIFAR dataset requires almost 100 days of computing time.

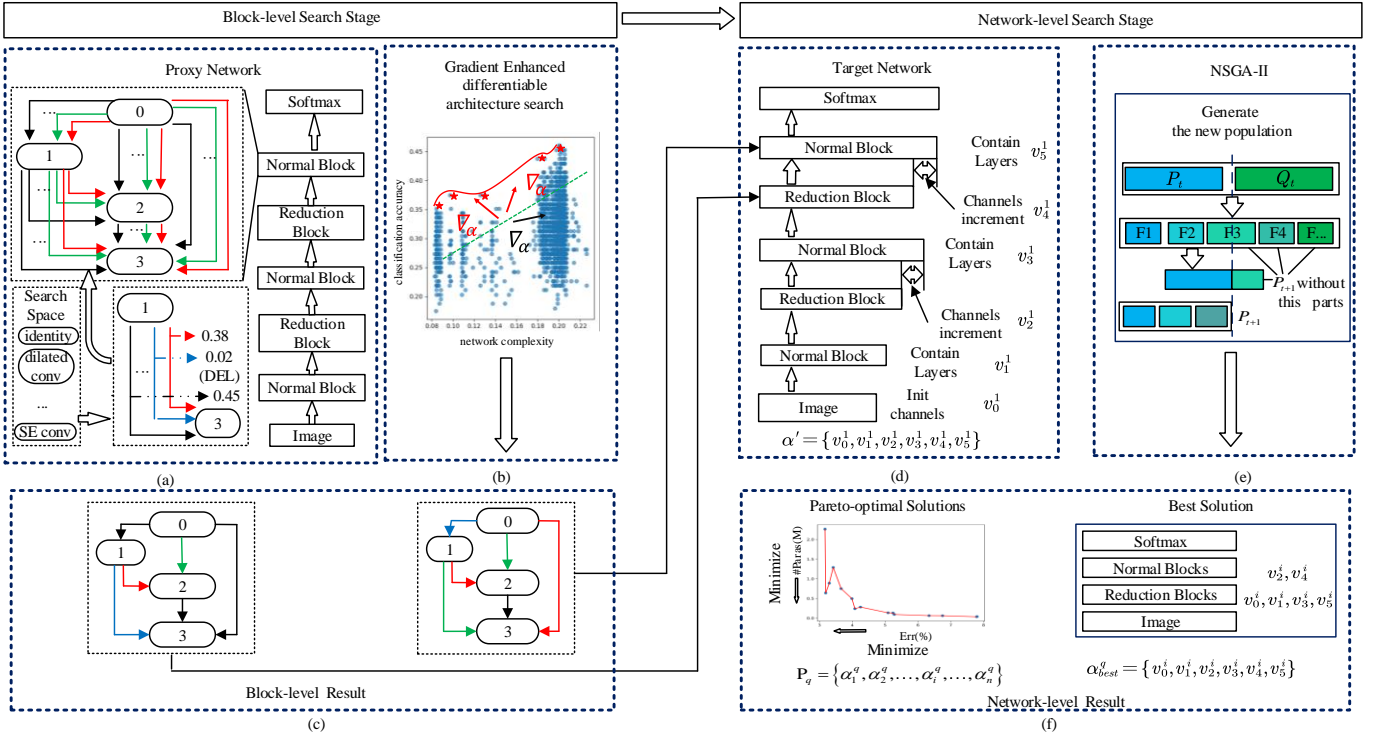

Fig.1. Schematic of the EG-DARTS search process. (a) A proxy network is composed of convolution operations and stacked blocks. (b) The method of gradient enhanced differentiable architecture search to maximize network classification accuracy and minimize network complexity. (c) Normal block and reduction block obtained from the block-level search. (d) The target network is constructed based on block parameterization (e) Multi-objective evolutionary optimization approach. (f) The Pareto solution set $P_q$ with diverse computational complexity is constructed by the proposed approach.

## III. PROPOSED METHOD

In this section, the details of the proposed framework are presented. We first give an overview of the proposed EG-DARTS. Then the design of the search space is given. Next, the core differentiable gradient-enhanced architecture search approach to construct the blocks is described. Finally, we combine the blocks with the multi-objective evolutionary architecture search algorithm to build the target network.

### A. Algorithm Overview

The actual application of NAS not only considers the objective of maximizing performance but also evaluates it against the conflicting objective in deployment scenarios within constrained computing resources. Therefore, our proposed algorithm will design high-performance neural network architectures with diverse computational complexity for resource-constrained scenarios. We regard the task as a multi-objective bilevel optimization problem, which can be expressed mathematically as:

$$\min F(\alpha) = (\mathcal{L}_{val}(\omega^*(\alpha),\alpha), \mathcal{C}(\alpha))^T$$
$$s.t. \ \omega^*(\alpha) = \arg\min_{\omega \in \Omega} \mathcal{L}_{train}(\omega,\alpha) \quad (1)$$

where $\alpha$ is the network architecture parameters and $\omega$ is the weights of the network. The lower-level objective $\mathcal{L}_{train}(\omega,\alpha)$ is the cross-entropy loss function on the training data. $\omega^*(\alpha)$ is the weights of the network where $\mathcal{L}_{train}(\omega,\alpha)$ is maximized. The upper-level objective $F(\alpha)$ consists of the loss function $\mathcal{L}_{val}(\omega^*(\alpha),\alpha)$ on the validation data and the computational complexity $\mathcal{C}(\alpha)$ of the network architecture. The goal for the multi-objective bilevel architecture optimization is to find $\alpha^\#$ that minimizes the validation loss $\mathcal{L}_{val}(\omega^\#(\alpha^\#),\alpha^\#)$ and the computational complexity $\mathcal{C}(\alpha^\#)$.

The purpose of the proposed EG-DARTS is to design high-performance architecture with diverse complexities for different specific scenarios. EG-DARTS consists of two stages and its schematic diagram is shown in Fig 1. EG-DARTS starts with the stage that we call the "block-level search stage", which is to search the optimal blocks from the search space. At this stage, as shown in Figs. 1(a)-(c), the block is designed by the gradient-based method, which is based on the progressive differentiable architecture search approach [11] and cooperates with architecture compression. As shown in Fig. 1(c), the reduction block and the normal block are searched in the proxy network. Likewise, another stage is called the "network-level search stage", which is to optimize the target network composed of obtained blocks. As shown in Figs. 1(e)-(f), the multi-objective evolutionary optimization method is combined with the objective of high performance and low complexity, and the target network is constructed based on the blocks obtained from the proxy network. Finally, the Pareto optimal solution $P_q = \{\alpha_1^q, \alpha_2^q, ..., \alpha_i^q, ..., \alpha_n^q\}$ with diverse computational complexity is constructed using NSGA-II as Fig. 1(f). In the remainder of this section (subsections B-D), a detailed description of the aforementioned components will be provided.

*B. Search Space and Encoding*

Neural architecture search is to design the optimal network architecture from different search spaces. The versatility of the search space has a major impact on the quality of the optimal architecture. To better design the search space, we need to understand the basic composition of neural networks. In the modern classic CNN, its architecture is mostly composed of block-level design and network-level design. Block-level design refers to hierarchical connection methods and calculation operations. Correspondingly, network-level design refers to changes in width, depth, and spatial resolution. Many manually-designed networks are based on this idea, such as ResNet [3] and Inception [2]. Similarly, various NAS methods are also searching for the best blocks. A block is a small module in CNN that is repeated multiple times to form the entire neural network. Therefore, choosing the search space should be oriented to block-level design. In this work, we leverage the search apace of PDARTS [21] as the baseline framework. In order to build a scalable network architecture, there are two types of blocks we build: (1) normal block, which returns feature maps with the same spatial resolution; and (2) reduction block, which returns feature maps with spatial resolution halved by a stride of two.

We search structure on the proxy network with fewer building blocks. The goal of designing the search space is to obtain the blocks with low computational complexity and high performance. The block is defined as a directed acyclic graph (DAG) of $N$ nodes and each node $x_i$ is a candidate representative of feature mapping in the network. The search space is denoted as $\mathcal{O}$, where each operation represents a candidate function $o(\cdot)$. Similarly, each directed edge $(i,j)$ is the candidate operation $o(x_i)$. Likewise, the block can be assumed to obtain a two-input node and a single output node. The input nodes are defined as block outputs in the previous two blocks. Therefore, an edge $(i,j)$ consists of a series of operations weighted $\alpha^{(i,j)}$ by architecture parameters, and it can be formulated as

$$f_{i,j}(x_i) = \sum_{o \in \mathcal{O}_{(i,j)}} \frac{\exp(\alpha_o^{(i,j)})}{\sum_{o' \in \mathcal{O}} \exp(\alpha_{o'}^{(i,j)})} o(x_i) \quad (2)$$

, where $i < j$, and the intermediate node is $x_j = \sum_{i<j} f_{i,j}(x_i)$. Thus, the output node is $x_{N-1} = concat(x_2, x_3, ..., x_{N-2})$, where $conact(\cdot)$ represents that all input signals in the channel dimension are concatenated. For the candidate convolution operations, we choose the computational operations between nodes from the following options, which are collected based on the excellent performance in the study of convolution neural networks.

- Identity (skip-connect)
- $3 \times 3$ max pooling (max_pool_3x3)
- $3 \times 3$ average pooling (avg_pool_3x3)
- efficient channel attention (EcaNet_3x3)
- $3 \times 3$ depthwise-separable convolution (sep_conv_3x3)
- $5 \times 5$ depthwise-separable convolution (sep_conv_5x5)
- $7 \times 7$ depthwise-separable convolution (sep_conv_7x7)
- $3 \times 3$ dilated convolution (dil_conv_3x3)
- $5 \times 5$ dilated convolution (dil_conv_5x5)
- $3 \times 3$ local binary convolution (Lbcnn_3x3)
- $5 \times 5$ local binary convolution (Lbcnn_5x5)
- $1 \times 7$ then $7 \times 1$ convolution (conv_7x1_1x7)

As shown in Fig. 1(a), during the algorithm search initialization at the first stage, the connection between two nodes is the operations of the search space described above, and there are 12 operations between two nodes. By the method of gradient enhanced differentiable architecture search, the result block contains only one computational operation per two nodes.

*C. Blocks-Level Search Strategy*

The purpose of the search at the first stage is to obtain the normal blocks and reduction blocks. The task of architecture search in the proposed gradient-based method can be understood as designing blocks with high performance and low complexity from the search space $\mathcal{O}$, through learning a set of continuous variables $\alpha^* = \{\alpha^{(i,j)}\}$. According to Eq. (1), the discrete architecture can be obtained by pruning the mixed computational operations, i.e., $o^{(i,j)} = \arg\max_{o \in \mathcal{O}} \alpha_o^{(i,j)}$. It is difficult to optimize network architecture performance and complexity based on gradients. To simplify the process, we first minimize validation loss $\mathcal{L}_{val}(\omega^*(\alpha), \alpha)$ from Eq. (1) and it can be simplified as:

$$\mathcal{L}_{val}(\omega^*(\alpha), \alpha) \approx \nabla_\alpha \mathcal{L}_{val}(\omega - \varepsilon \nabla_\omega \mathcal{L}_{train}(\omega, \alpha), \alpha) \quad (3)$$

where $\omega$ represents the weights of the current network and $\varepsilon$ represents the learning rate at each step of the architecture optimization process. Then, Eq. (3) can be expended by the chain rule as:

$$\nabla_\alpha \mathcal{L}_{val}(\omega', \alpha) - \varepsilon \frac{\nabla_\alpha \mathcal{L}_{train}(\omega^+, \alpha) - \nabla_\alpha \mathcal{L}_{train}(\omega^-, \alpha)}{2\epsilon} \quad (4)$$

where $\epsilon = 0.01/|\nabla_{\omega'} \mathcal{L}_{val}(\omega', \alpha)|_2$ represents a small scalar and $\omega^\pm = \omega \pm \epsilon \nabla_{\omega'} \mathcal{L}_{val}(\omega', \alpha)$. $w' = w - \xi \nabla_w \mathcal{L}_{train}(w, \alpha)$ is the weights of the one-step forward network. Since Eq. (4) simplifies the architecture search algorithm, the complexity of NAS is reduced from $O(|\alpha||\omega|)$ to $O(|\alpha| + |\omega|)$.

In order to minimize the classification loss and minimize the complexity of the network architecture, we propose an improvement to the gradient $\nabla_\alpha$ updating of the architecture parameters $\alpha^{(i,j)}$ as shown in Fig. 1(b). The relationship between classification loss and the network complexity is assumed to be approximated as:

$$y = h_\theta(x) = \theta_0 + \theta_1 x \quad (5)$$

where $x$ and $y$ represent network complexity and classification loss respectively. Correspondingly, $\theta_0$ can be estimated by the least square method as:

$$\theta_1 = \frac{m\sum_{i=1}^{m}(x^i y^i) - \sum_{i=1}^{m} x^i \sum_{i=1}^{m} y^i}{m\sum_{i=1}^{m}(x^i)^2 - \left(\sum_{i=1}^{m} x^i\right)^2} \quad (6)$$

where $(x^i, y^i) \in \{(x^1, y^1), ..., (x^m, y^m)\}$. The red $\nabla_\alpha$ in Fig. 1(b) indicates the gradient in the direction of low network complexity and high classification loss of the network. The black $\nabla_\alpha$ indicates the gradient in the direction of high network complexity and high classification loss. Therefore, red $\nabla_\alpha$ is the required gradient, and black $\nabla_\alpha$ is the unnecessary gradient. To distinguish between red $\nabla_\alpha$ and black $\nabla_\alpha$, a linear relationship between the network complexity and classification loss is fitted by least square. In the process of updating the architecture parameter gradient $\nabla_\alpha$, the relationship $\theta_1$ between the network complexity and the classification loss can be expressed as:

$$\nabla_\theta = \frac{x^k - x^{k-1}}{y^k - y^{k-1}} \quad (7)$$

where $x^k$ and $x^{k-1}$ represents the network complexity at the $k$th and the $(k-1)$th step respectively. Similarly, $y^k$ and $y^{k-1}$ represent the classification loss at the $k$th step and the $(k-1)$th step. When $\nabla_\theta > \theta_1$ or $\nabla_\theta > 0$, the gradient $\nabla_\alpha$ is judged to be the red gradient. When $\nabla_\theta < \theta_1$ the gradient $\nabla_\alpha$ is judged to be the black gradient.

To simultaneously minimize classification loss and network complexity, the red gradient should be enhanced. Thus, a variable coefficient $(1 + \Phi(\alpha))$ is added to Eq. (4) as:

$$\nabla'_\alpha \mathcal{L}_{val}(\omega', \alpha) - \varepsilon \frac{\nabla'_\alpha \mathcal{L}_{train}(\omega^+, \alpha) - \nabla'_\alpha \mathcal{L}_{train}(\omega^-, \alpha)}{2\epsilon} \quad (8)$$

where $\nabla'_\alpha = \nabla_\alpha (1 + \Phi(\alpha))$, and $\Phi(\alpha)$ is defined as:

$$\Phi(\alpha) = \sigma \frac{|\nabla_\theta|}{\theta_1} \quad (9)$$

In Eq. (9), $\sigma$ represents the indicating function and is represented as:

$$\sigma = \begin{cases} 1 & \text{if } \nabla_\theta > \theta_1 \text{ or } \nabla_\theta > 0 \\ 0 & \text{if } \nabla_\theta < \theta_1 \end{cases} \quad (10)$$

As presented in Eqs. (5)-(10), when updating the architecture parameters $\alpha^{(i,j)}$, the enhancement of the architecture gradient $\nabla_\alpha$ reduces the network complexity and minimizes the classification loss.

*D. Network-Level Search Strategy*

Through the first stage (blocks-level search stage), we have obtained the normal block and reduction block. However, the shared blocks need to be optimized because of the gap between the proxy network and target network. Therefore, when building block-based target networks based on blocks, a multi-objective evolutionary optimization algorithm is used. The depth and width of the target network will be the search space at this stage. The detailed configuration of the search space in this stage will be introduced in Section IV-A. For the design of the spatial resolution in the target network, we adopt the one used in [20, 32].

**Algorithm 1** Multi-Objective Evolutionary Optimization Architecture

Input: The population size $N$, the maximal generation number $T$, the crossover probability $\mu$, and the mutation probability $\nu$.
1. Randomly initialize the population $P$ with the size of $N$.
2. Compute the network complexity $\mathcal{C}(P)$ as Eq.1 and compute the classification accuracy $\mathcal{L}_{train}(P)$ as [41]
3. $[F_1, F_2, ...] \leftarrow$ Fast non-dominated sorting $(\mathcal{L}_{train}(P), \mathcal{C}(P))$
4. $[G_1, G_2, ...] \leftarrow$ Crowding distance sorting $[F_1, F_2, ...]$
5. **For** $t = 1, 2, 3, ..., T$ **do**
6.    Clear offspring population $Q \leftarrow \varnothing$, $i \leftarrow 0$
7.    **While** $i < N$ **do**
8.       $p \leftarrow (P, [G_1, G_2, ...], [F_1, F_2, ...])$ select individual by the tournament.
9.       $q \leftarrow Crossover(p, \mu)$
10.      $q \leftarrow Mutatuin(p, \nu)$
11.      $Q \leftarrow Q \cup q; i = i + 1$
12.    **End While**
13.    Compute the network complexity $\mathcal{C}(Q)$ as Eq.1 and compute the classification accuracy $\mathcal{L}_{train}(Q)$ as [41]
14.    $[F_1, F_2, ...] \leftarrow$ Fast non-dominated sorting $(\mathcal{L}_{train}(Q), \mathcal{C}(Q))$
15.    $[G_1, G_2, ...] \leftarrow$ Crowding distance sorting $[F_1, F_2, ...]$
16.    $P \leftarrow (P \cup Q, [G_1, G_2, ...], [F_1, F_2, ...])$ select individual by the tournament.
17.    $t = t + 1$
18. **End for**
Output: The Pareto population $P$

At the network-level search stage, the algorithm is an iterative process. In the search process, the initial architecture as a group is improved gradually. At each iteration, a set of offspring is created by applying the mutations and crossover operations. Fig. 1(d) shows an example of applying this algorithm into searching on CIFAR10. Because the resolution of CIFAR samples is $32 \times 32$, the target network consists of two down-sampling layers and the final stage of the target network is a fully connected layer. As the example of searching architecture for CIFAR10, the search space of multi-objective evolutionary algorithm contains six variables. We use a mixed encoding of real and integer variables to accommodate the construction of a target network. Integer variables $v_1^1, v_3^1, v_5^1$ represent the size of the layers with normal blocks between reduction blocks. Real variables $v_2^1, v_4^1$ represent the increment

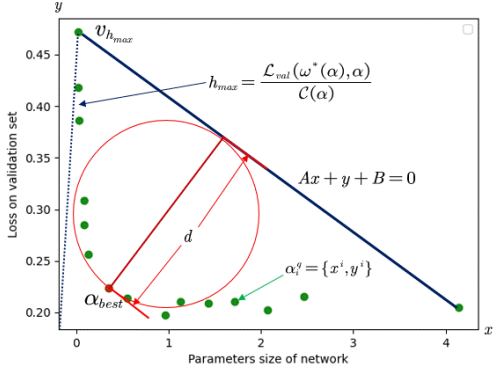

Fig.2. Schematic of the decision-making method for the optimal neural network architecture. The green points represent individuals in the Pareto solution set. By calculating the distance of each individual to the linear function $Ax+y+B=0$, the optimal individual is obtained.

---

**Algorithm 2** Decision-Making Approach for the Optimal Neural Network Architecture
---
**Input**: Pareto solution set $P_q = \{\alpha_1^q, \alpha_2^q, ..., \alpha_i^q, ..., \alpha_n^q\}$, where $\alpha_i^q = \{x^i, y^i\}$.
1. Based on the point $(x_0^{max}, y_0)$ with the largest parameter size of the network architecture and the point $(x_1^{min}, y_1)$ with the smallest parameter size, the linear function $Ax+y+B=0$ is constructed, where $A = -\dfrac{y_0 - y_1}{x_0^{min} - x_1}$ and $B = -\dfrac{y_0 - y_1}{x_0^{max} - x_1^{min}} x_0 - y_0$.
2. $S_d \leftarrow \varnothing$
3. **For** $i = 1, 2, 3, ..., n$ **do**
4. $\quad$ Calculate the distance $d_i = \dfrac{Ax^i + y^i + B}{\sqrt{A^2 + 1}}$
5. $\quad$ $S_d \leftarrow S_d \cup d_i$
6. **End for**
7. Get the $\alpha_{best} = \arg\max\limits_{\alpha(x^i, y^i)}(S_d)$

**Output:** The optimal individual $\alpha_{best}$.

---

of the channels when the feature is down-sampling. The integer variable $v_0^1$ indicates the size of the first convolutional layer channels.

The purpose of optimizing the architecture is to obtain the network architecture with low complexity and excellent performance. However, if we use $h = \dfrac{\mathcal{L}_{val}(\omega^*(\alpha), \alpha)}{\mathcal{C}(\alpha)}$ to measure the value of individuals in the Pareto solution set, the individual $v_{h_{max}}$ with the maximum value $h$ cannot achieve the goal of optimizing the architecture. Instead, we need the best individual as $\alpha_{best}$ in Fig. 2. Thus, a decision-making method is designed to obtain the optimal neural network architecture.

The proposed decision-making method is implemented by measuring the distance $d$ from the individual to the linear function $Ax+y+B=0$, and the optimal individual is the one with the largest distance $d_{max}$. The specific algorithm is shown in Algorithm 2. The optimal individual selected by this method has taken into account the performance and complexity of the neural network architecture simultaneously.

## IV. EXPERIMENTAL SETUP AND RESULT

In this section, we first present the implementation details of our experiments. Then we discuss the effectiveness of the block-level search process and the properties of the Pareto solution set. Next, we analyze the training process of block-level search and network-level search on CIFAR10 and CIFAR100. Finally, we analyze ablation experiments about the parameter settings of the proposed framework and the effectiveness of algorithm components.

### A. Implementation Detail

In order to demonstrate the effectiveness of the proposed algorithm, we compare the Pareto solution set of EG-DARTS with those produced by other state-of-the-art methods. Based on the ease of use and efficiency of algorithm comparison, these methods are compared on the CIFAR datasets. All comparison experiments we set up are completed on CIFAR10 and CIFAR100.

Moreover, the comparison methods we select are mainly categorized into two groups: the neural network architectures designed by human experts and NAS-based. The architectures designed by human experts include ResNet, DenseNet, and DPN, etc. Another group consists of earlier NAS methods and more recent methods based on EA or gradient-based, which have been proved to be effective in completing various computer version tasks. The NAS method we have proposed is to address multi-objective network architecture optimization, namely, maximizing classification accuracy and minimizing the computational complexity of the network. In the literature, there are various metrics that can represent the complexity of the network, including the parameter size, the inference time, and the number of FLOPS. Due to the difference and inconsistency of factors such as computing environment and ambient temperature, the inference time may not be reliably used as a network complexity metric. Therefore, we choose parameters size and FLOPS as the indicators of network complexity.

In the process of the block-level stage, we use the standard Stochastic Gradient Descent (SGD) method to learn the weights of the network and the weights of the architecture parameters. In the network-level search process, the SGD is used to estimate the classification accuracy of each corresponding to the network architecture. At the same time, NSGA-II is used to design low-complexity and high-performance target networks. To make the search process more efficient and meet the search requirements, the computational resource consumption of the algorithm needs to be reduced as much as possible. Table I summarizes the hyper-parameter settings in EG-DARTS related to search space, gradient descent training, and search strategy. All the experiments are conducted on four GPUs with NVidia GeForce RTX TITAN and two CPUs with Intel Xeon Silver 4214. The codes of EG-DARTS are coded by Python 3.6.9 and Pytorch 1.7.

To verify the effectiveness of the proposed EG-DARTS approach, a series of experiments are designed and performed. The designed experiments can be divided into two different cat-egories. The first category is comparison experiments. The state-of-the-art neural network architectures that are hand-crafted such as VGG, ResNet, and DenseNet, are selected to compare with EG-DARTS. The neural network architecture search algorithms such as AE-CNN and Genetic CNN are also chosen for comparison. The second experiment category covers the performance analyses of different parameters and components of EG-DARTS and further explanation about the impact of some important hyper-parameters on the performance with designed neural networks.

TABLE I: A SUMMARY OF HYPER-PARAMETER SETTING.

| Categories | Parameter name | Parameter value |
|---|---|---|
| Gradient Descent (Block-Level) | batch size | 96 |
| | weight decay | 5.00E-04 |
| | training epochs | 40 |
| | learning rate | 0.025 |
| Gradient Descent (Network-Level) | batch size | 128 |
| | weight decay | 5.00E-04 |
| | training epochs | 36/600 |
| | learning rate | 0.025 with Cosine Annealing |
| Search Space (Network-Level) | $v_0$ | (8,60) |
| | $v_1$ | (1,6) |
| | $v_2$ | (1,6) |
| | $v_3$ | (1,6) |
| | $v_4$ | (1,3) |
| | $v_5$ | (1,3) |
| Search Strategy (Network-Level) | population size | 15 |
| | generations | 20 |
| | crossover probability | 0.9 |
| | mutation probability | 0.1 |

*B. Efficiency of EG-DARTS*

In the literature, the parameter size of the network architecture is used to measure the complexity of the network architecture for most of neural architecture searching approaches based on the CIFAR dataset. Here, to prevent potential discrepancies from reimplementation, we also use the parameter size to compare the complexity of the network. In addition, to facilitate comparison with other peer algorithms, we follow the training process presented in [21]. Table Ⅱ shows the performance of the proposed method, where the results are obtained by training the network architecture using EG-DARTS on both CIFAR10 and CIFAR100. The method proposed in this work has obvious performance advantages compared with other peer methods considered. Specifically, on the CIFAR10, the accuracy of our algorithm is slightly lower than that of DARTS [21], while the efficiency of network parameter size is five times more efficient than that of peers. Furthermore, EG-DARTS achieves a network architecture that is three times fewer parameter size than DARTS [21] to obtain an error rate of less than 20% on CIFAR100. However, we find that transplanting the architectures searched on CIFAR10 to the CIFAR100 dataset only yields a classification accuracy of $21.78\%$, and transplanting the architecture searched on CIFAR100 to the CIFAR10 dataset yields a classification test accuracy of $3.80\%$.

*C. Block-level Search Analysis*

The block topology can be obtained using the approach of enhancing gradient by searching in a proxy network, which is consisted of the initial stage, the intermediate stage, and the final stage. At the initial stage, the proxy network depth is five, and there are twelve operations between two nodes. According to the order of the architecture parameter $\alpha$, four operations with the smallest $\alpha$ are deleted. At the intermediate stage, the depth of the trained proxy network is eleven, and there are eight operations between two nodes. Once this stage is completed, only four operations are left between the nodes. At the final stage, the depth of the trained proxy network is seventeen, and there are four operations between two nodes. What we can obtain from the end-stage is the final block topology where each node is connected by a remaining operation and the other three operations are deleted with the minimum three architecture parameters $\alpha$. The block topology diagram obtained from the block-level search is shown in Fig. 3.

Figs. 3(a)-(b) show the block topology searched on CIFAR10, where Fig. 3(a) is the structure of the normal block, and Fig. 3(b) is the structure of the reduction block. The resolution of the feature map remains unchanged in the normal block but the resolution is reduced to a half in the reduction block. Therefore, there will be a large number of down-sampling operations between nodes in the reduction block. Most operations in the normal block are convolution ones to keep the resolution of the data feature map unchanged while further extracting the features of the data information. There are more sep_conv_3x3 operations in these blocks as demonstrated in Figs. 3(e)-(f), compared with DARTS [20]. On the other hand, the block structure Figs. 3(a)-(b) obtained by our approach contains more EcaNet_3x3 operations. It has been revealed in references [6] and [42] that the main purpose of the sep_conv_3x3 operation is to reduce the size of parameters, and the EcaNet_3x3 operation is a module of the attention mechanism, which reduces operation parameter size while obtaining better feature extraction capabilities. Therefore, the Eca_Net operations are the main part of the blocks. Figs. 3(c)-(d) demonstrate the search results on CIFAR100. Compared with the search results on CIFAR10, skip-connect operations appear more times in the block structure of Figs. 3(c)-(d). The reason for the block structure on CIFAR100 may be related to its data distribution. Compared with 6000 images in each class of CIFAR10, there are only 600 images in each class of CIFAR100. The skip-connection in the blocks is to avoid network overfitting.

*D. Network-level Search Analysis*

Based on the blocks obtained by the proxy network search, the target network is optimized at the network-level stage. Table Ⅲ shows the Pareto solution set on CIFAR10 and CIFAR100. Architecture parameters $v_0$-$v_5$ need to be searched in the target network, as shown in Fig. 1(d). Table III demonstrates the composition of $v_0$-$v_5$ is based on the Pareto solution set. The channel variable $v_0$ is close to 50, and the layers $v_1$ of normal blocks between reduction blocks is 6. Equally, $v_0$ represents the number of convolution kernel channels at the input part of the network. The feature map in the

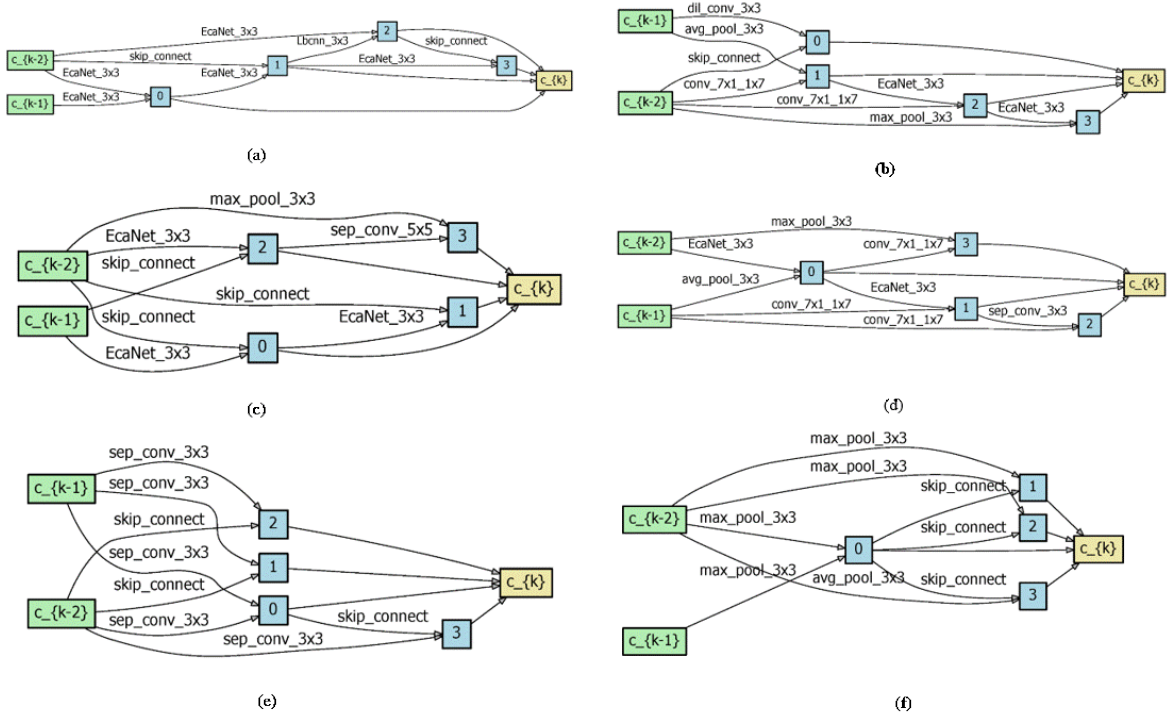

Fig. 3 Block architecture learned on different classification tasks. (a) and (b) represent the normal block and the reduction block, respectively learned on CIFAR10 by EG-DARTS. (c) and (d) represent the normal block and the reduction block, respectively learned on CIFAR100 by EG-DARTS. (e) and (f) represent the normal block and the reduction block, respectively from [20] on CIFAR10.

TABLE II. COMPARISON OF IMAGE CLASSIFICATION ARCHITECTURES ON CIFAR10 AND CIFAR100.

| Architecture | Test Err. (%) | | Parameters | GPU Days | Search Method |
|---|---|---|---|---|---|
| | CIFAR10 | CIFAR100 | | | |
| DenseNet-BC [43] | 3.46 | 17.18 | 26.6 | - | manual |
| ResNet [3] | 4.61 | 22.10 | 1.7 | - | manual |
| NASNet-A [30] + curout | 2.65 | - | 3.3 | 1800 | RL |
| AmoebaNet-A [44] + curout | 3.34 | - | 3.2 | 3150 | EA |
| AmoebaNet-B [44] + curout | 2.55 | - | 2.8 | 3150 | EA |
| Hireachical Evolution [45] | 3.75 | - | 15.7 | 300 | EA |
| Genetic CNN [28] | 7.10 | 29.05 | - | 14 | EA |
| Block-QNN-S [31] | 4.38 | 20.65 | 6.1 | 90 | RL |
| CGP-CNN [26] | 5.98 | - | 2.64 | 27 | EA |
| PNAS [46] | 3.41 | | 3.2 | 255 | SMBO |
| ENAS [47]+ cutout | 2.89 | | 4.6 | 0.5 | RL |
| DARTS (first order) [20] + cutout | 3.00 | 17.76 | 3.3 | 1.5 | gradient-based |
| DARTS (second order) [20] + cutout | 2.76 | 17.54 | 3.3 | 4 | gradient-based |
| EG-DARTS(On CIFAR10) | 3.19 | 21.78 | 0.64 | 7 | gradient-based + EA |
| EG-DARTS(On CIFAR100) | 3.80 | 19.72 | 0.97 | 7 | gradient-based + EA |

input part of the network contains substantial original information. This suggests that more convolution kernel channels and convolution layers can help the network filter the noise of the data and extract the main features. The values of $v_2$ and $v_3$ are also small, as that this part of the feature maps already contains the main characteristics of the data. For this reason, only a few network parameters are needed to fit the distribution of the data. In TableIII(a), individuals are ranked according to the parameter size. From Table III(a), we can find that the attributes of the optimal individual are significantly different from those of other individuals. For example, the optimal individual obtained from the decision-making method in Section III-D is presented in the fifth line of Table III(a). Compared with the individual in the fourth row of Table III(a), the optimal individual has fewer network parameters and higher classification performance, but the optimal individual leads to higher FLOPS and longer inference time. Based on the value of $v_0$ - $v_5$, the reason for the longer inference time is the target network contains more block layers. For this reason, the optimal individual based on CIFAR100 contains the most block layers, as shown in Table III (b), and it also has the longest inference time. This suggests that the complexity of the network is not linearly related to the inference time. Next, we analyze the correlation between inference time and network architecture,

TABLE III (A) THE PARETO SOLUTION SETS SEARCHED ON CIFAR10 BASED ON THE BLOCKS OF FIGS. 3(A)-(B). (B) THE PARETO SOLUTION SET SEARCHED BY THE APPROACH ON CIFAR100 BASED ON THE BLOCKS OF FIGS. 3(C)-(D). #PARAMS REPRESENTS THE PARAMETER SIZE OF THE NETWORK. ERR. REPRESENTS THE CLASSIFICATION ERROR RATE OF THE NETWORK. FLOPS IS THE NUMBER OF FLOATING-POINT OPERATIONS OF THE NETWORK. LATENCY REPRESENTS THE AVERAGE INFERENCE TIME CONSUMED BY A SINGLE IMAGE INFERENCE OBTAINED BY REPEATING IMAGE INFERENCE 1000 TIMES ON THE COMPUTER WITH A GPU OF 1080TI. DEPTH IS THE NUMBER OF CNN BLOCK LAYERS.

(a) CIFAR10

| $v_0$ | $v_1$ | $v_2$ | $v_3$ | $v_4$ | $v_5$ | #Params | Err. (%) | FLOPS (M) | Latency (ms) | Depth |
|---|---|---|---|---|---|---|---|---|---|---|
| 50 | 6 | 2 | 2 | 2.50 | 1.52 | 2.26 | 3.17 | 541.161 | 25.83 | 10 |
| 35 | 4 | 3 | 2 | 2.50 | 1.70 | 1.29 | 3.42 | 259.59 | 23.65 | 9 |
| 35 | 4 | 3 | 2 | 2.50 | 1.16 | 0.89 | 3.3 | 228.38 | 23.65 | 9 |
| 27 | 2 | 3 | 2 | 2.42 | 1.79 | 0.75 | 3.66 | 130.72 | 19.41 | 7 |
| **50** | **6** | **2** | **3** | **1.20** | **1.09** | **0.64** | **3.19** | **305.32** | **27.92** | **11** |
| 27 | 4 | 3 | 2 | 1.36 | 2.84 | 0.50 | 3.99 | 99.56 | 23.64 | 9 |
| 28 | 2 | 2 | 2 | 1.31 | 1.83 | 0.28 | 4.25 | 64.44 | 17.32 | 6 |
| 28 | 2 | 2 | 2 | 1.20 | 1.83 | 0.24 | 4.08 | 58.27 | 17.3 | 6 |
| 16 | 2 | 3 | 3 | 1.87 | 1.09 | 0.14 | 5.09 | 32.45 | 21.54 | 8 |
| 11 | 3 | 3 | 2 | 2.50 | 1.68 | 0.13 | 5.24 | 25.64 | 21.53 | 8 |
| 9 | 6 | 4 | 4 | 2.33 | 1.28 | 0.09 | 5.28 | 22.63 | 34.47 | 14 |
| 9 | 6 | 4 | 4 | 1.85 | 1.28 | 0.06 | 6.34 | 18.04 | 34.44 | 14 |
| 15 | 2 | 2 | 3 | 1.19 | 1.28 | 0.06 | 6.75 | 16.83 | 19.42 | 7 |
| 9 | 6 | 4 | 4 | 1.41 | 1.19 | 0.04 | 7.8 | 15.01 | 34.44 | 14 |
| 9 | 6 | 4 | 2 | 1.41 | 1.16 | 0.03 | 7.79 | 14.72 | 30.12 | 12 |

(b) CIFAR100

| $v_0$ | $v_1$ | $v_2$ | $v_3$ | $v_4$ | $v_5$ | #Params | Err. (%) | FLOPS (M) | Latency (ms) | Depth |
|---|---|---|---|---|---|---|---|---|---|---|
| 54 | 6 | 2 | 2 | 2.93 | 2.42 | 4.14 | 20.50 | 679.43 | 13.60 | 10 |
| 54 | 6 | 2 | 2 | 2.80 | 1.76 | 2.47 | 21.54 | 531.07 | 13.22 | 10 |
| 46 | 6 | 2 | 2 | 2.41 | 2.38 | 2.07 | 20.22 | 383.32 | 13.21 | 10 |
| 51 | 6 | 2 | 2 | 2.61 | 1.61 | 1.72 | 21.09 | 409.62 | 13.22 | 10 |
| 56 | 6 | 2 | 3 | 2.29 | 1.15 | 1.43 | 20.90 | 417.36 | 14.27 | 11 |
| 56 | 6 | 2 | 2 | 2.29 | 1.03 | 1.12 | 21.07 | 395.41 | 13.19 | 10 |
| **48** | **6** | **3** | **4** | **1.12** | **2.38** | **0.97** | **19.72** | **244.62** | **16.26** | **13** |
| 44 | 6 | 3 | 2 | 1.08 | 2.46 | 0.55 | 22.42 | 188.23 | 14.24 | 11 |
| 35 | 6 | 3 | 2 | 1.07 | 2.45 | 0.35 | 23.40 | 116.16 | 14.22 | 11 |
| 22 | 3 | 6 | 2 | 1.17 | 1.79 | 0.13 | 25.62 | 37.18 | 14.25 | 11 |
| 14 | 2 | 2 | 3 | 1.76 | 1.45 | 0.08 | 28.50 | 16.13 | 10.08 | 7 |
| 10 | 5 | 2 | 2 | 1.98 | 2.37 | 0.08 | 30.90 | 15.11 | 12.14 | 9 |
| 8 | 5 | 2 | 3 | 1.96 | 1.03 | 0.03 | 38.61 | 8.50 | 13.20 | 10 |
| 8 | 2 | 2 | 2 | 1.83 | 1.12 | 0.02 | 41.83 | 5.44 | 9.07 | 6 |
| 8 | 5 | 2 | 3 | 1.14 | 1.11 | 0.02 | 47.16 | 6.73 | 13.19 | 10 |

as shown in Fig. 4 and Fig. 5.

Fig. 4(a) shows the relationships among the classification error rate, network parameter size, and inference time of the Pareto solution set for CIFAR10, where the data of the inference time is more discretized. Fig. 4(b) shows the Pareto front obtained by EG-DARTS. Fig. 4(c) is consistent with Fig. 4(b), which shows that the network architecture attributes represented by FLOPS are similar to those represented by the parameter size of the network. Fig. 4(d) is quite different from that in Fig. 4(b), which further suggests that the inference time is not linearly related to the parameter size. Similarly, the above-mentioned data distribution also appears on CIFAR100, as shown in Figs. 5(a)-(d).

To further study which network architecture attributes affect the network inference time, we construct the trend curve based on network parameters size, network depth, and network inference time normalized data, as shown in Figs. 4(e)-(f) and Figs. 5(e)-(f). It can be concluded that the inference time of the network has the strongest correlation with the depth of the network, and the curve trends of the inference time and network depth are almost coincident. This suggests that when the data is calculated in the GPU, the inference time is mainly generated by the feature map through each layer of convolution. The channels of each layer affect the amount of data-parallel calculation and do not affect the inference time. To summarize, the parameters size of the network and the number of floating-point operations represent the same network attributes, and the inference time cannot be expressed by the parameters size and the number of floating-point operations. The inference time is linearly related to the depth of the network.

*E. Ablation Study*

Fig. 6 shows the Pareto solution set under different generations on both CIFAR10 and CIFAR100. It can be seen from Fig. 6(a) that the increase of generation on CIFAR10 has little effect on the approximate Pareto front boundary. This phenomenon indicates that the number of generations is set as 5 based on CIFAR10, EG-DARTS has converged. Fig. 6(b) is the ablation experimental result on CIFAR100. When the number of generations increases, the front boundary of Pareto does not change significantly. As described, the number of generations is set to 5, and it can ensure convergence on CIFAR10 and CIFAR100.

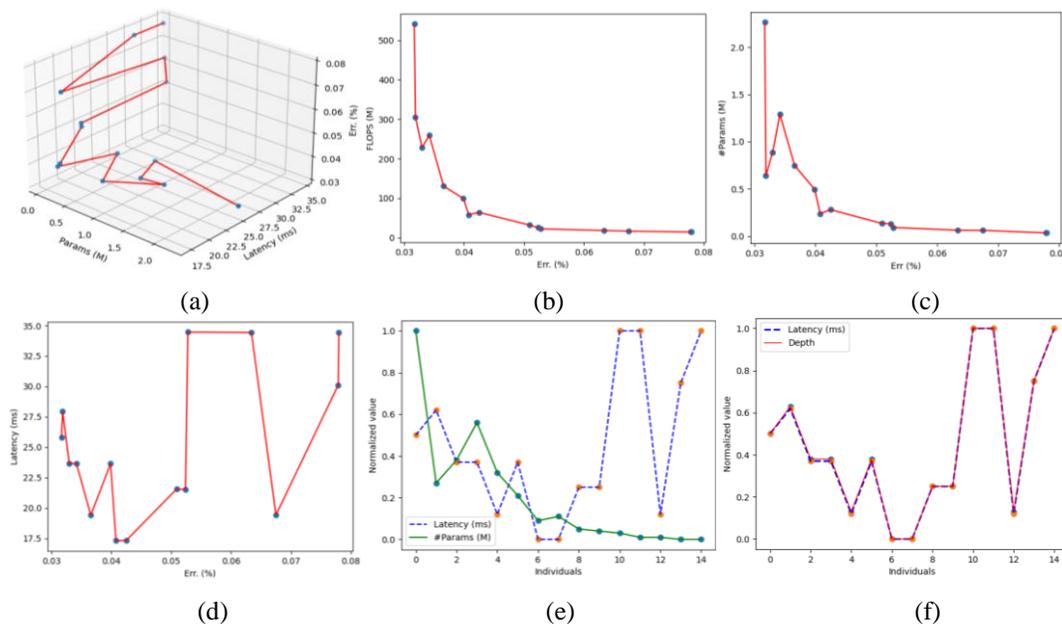

Fig. 4. The characteristic relationship diagram of the obtained Pareto solution set on CIFAR10. The diagram is designed to present the relationship between accuracy, network complexity, and inference time. (a) A 3-D diagram of the relationship between classification error rate (Err.), network parameter size (#Params), and inference time (Latency). (b) Pareto front with the classification error rate and the number of floating-point operations. (c) The relationship between the classification error rate and parameter size. (d) The relationship between the classification error rate and inference time. (e) The individuals of the Pareto solution set are normalized on the parameters size and the inference time. (f) The individuals of the Pareto solution set are normalized on the network depth and the inference time.

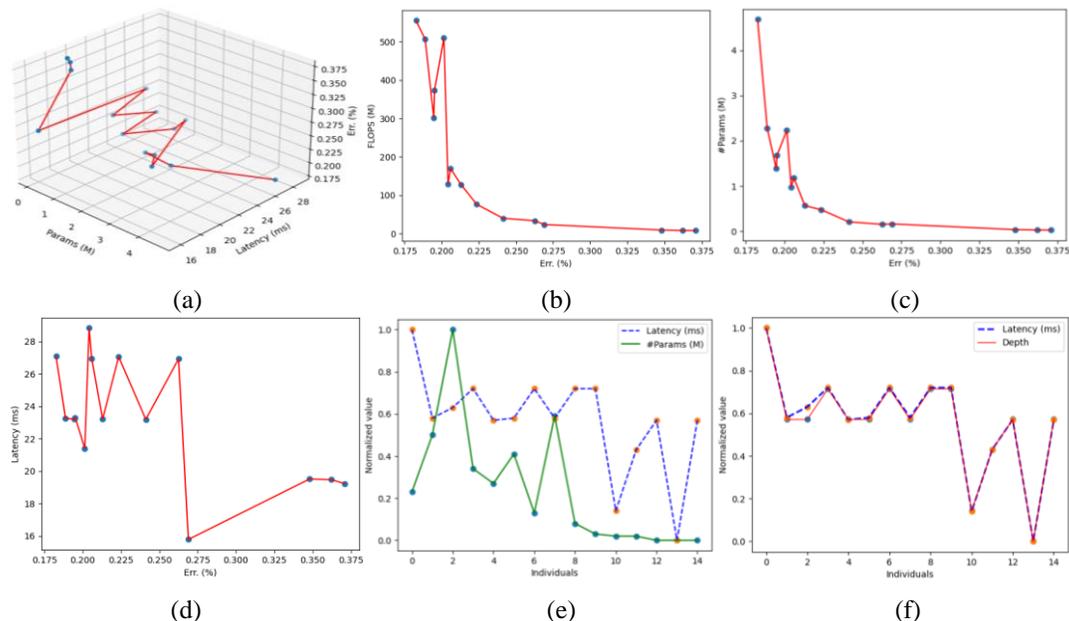

Fig. 5 The characteristic relationship diagram of the Pareto solution set on CIFAR100. (a) A 3-D diagram of the relationship among classification error rate, network parameter size, and inference time. (b) Pareto front with the classification error rate and the number of floating-point operations. (c) The diagram of the relationship between the classification error rate and parameter size. (d) The diagram of the relationship between the classification error rate and inference time. (e) The individuals of the Pareto solution set are normalized on the parameters size and the inference time. (f) The individuals of the Pareto solution set are normalized on the network depth and the inference time.

To verify the effectiveness of the block-level search in Section III-C, we apply the blocks from P-DARTS [21] for the network-level search of EG-DARTS and the results are shown in Table IV and Fig. 6. Table IV shows that the optimal target network constructed by blocks and searched by EG-DARTS has the same classification performance but fewer parameters. It can be concluded that the block-level approach of EG-DARTS can design blocks with high performance and low complexity. To judge the blocks of the target network are optimal, we use the block-level approach to directly optimize the blocks for the second time based on the optimal target network architecture parameters $v_0$ - $v_5$. The experimental results are shown in Table V. It can be seen from the experimental results that the classification performance of the target network with the second optimized blocks is lower than the target network with the blocks in Section IV-C. We can infer that the network architecture parameters $v_0$ - $v_5$ cause the target network to be too complex, and the second optimization of the blocks fails.

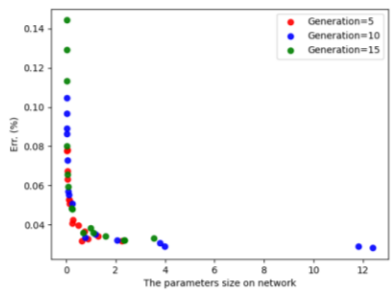

(a) CIFAR10

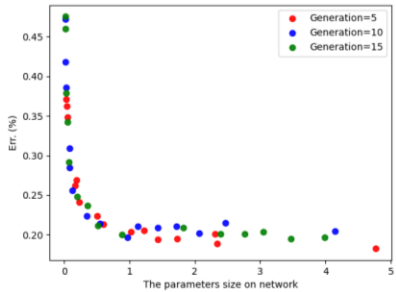

(b) CIFAR100

Fig. 5. Pareto front with the classification error rate and parameter size when the number of generations is set as 5, 10, and 15.

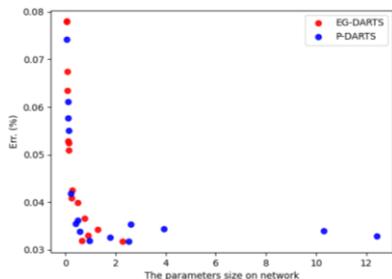

Fig. 6. Pareto front with the classification error rate and parameter size on the blocks through P-DARTS [21] and EG-DARTS.

TABLE IV. COMPARISON OF IMAGE CLASSIFICATION PERFORMANCE FOR TARGET NETWORK WHICH BLOCKS PRODUCED BY P-DARTS AND EG-DARTS.

| Block search method | #Para (M) | Err. (%) |
|---|---|---|
| P-DARTS Blocks | 0.96 | 3.19 |
| EG-DARTS Blocks | **0.64** | **3.19** |

TABLE V. COMPARISON OF IMAGE CLASSIFICATION PERFORMANCE FOR TARGET NETWORK BLOCKS PRODUCED THROUGH THE FIRST BLOCKS-LEVEL SEARCH AND THE SECOND BLOCKS-LEVEL SEARCH.

| Blocks-level search method | #Para (M) | Err. (%) |
|---|---|---|
| First blocks-level search | **0.64** | **3.19** |
| Second blocks-level search | 0.65 | 3.39 |

## V. CONCLUSIONS AND FUTURE WORK

In this paper, we present EG-DRATS, a multi-objective evolutionary algorithm for neural architecture searching. EG-DRATS has completely designed the network architecture at two stages through block-level search and network-level search. We develop the gradient-based relaxation method for designing blocks with high performance and low complexity by enhanced gradient. Furthermore, the proposed framework applies the evolutionary multi-objective algorithm to complete the automatic design from blocks to the target network. This method provides a reliable technique to optimize the neural networks to achieve high-performance deployment on resource-constrained devices. The algorithm is validated on CIFAR10 and CIFAR100. What is more, at the network-level search stage, we analyze the relationships among network parameters, FLOPS, and inference time. We find that the inference time is directly related to the network depth. Optimizing the network architecture from the three objectives of network parameter size, classification performance and reasoning time is our future research direction.

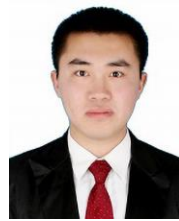

**Haichao Zhang** is a Ph.D. candidate at the College of Information Sciences and Technology, Donghua University, Shanghai, China. He obtained his M.S. degree in Textile Engineering from Inner Mongol University of Technology, Hohhot, China in 2016. His scientific interests include deep learning, machine vision, image processing, and artificial intelligence.

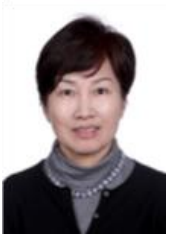

**Kuangrong Hao** is currently a Full Professor at the College of Information Sciences and Technology, Donghua University, Shanghai, China. She received her B.S. and M.S. degrees in mechanical engineering from Hebei University of Technology, Tianjin, China in 1984 and 1989, respectively. She received her M. S. and Ph.D. degrees in applied mathematics and computer sciences from Ecole Normale Supérieur de Cachan, France in 1991, and Ecole Nationale des Ponts et Chaussées, Paris, France in 1995, respectively. She has published more than 200 technical papers, and five research monographs. Her scientific interests include intelligent perception, intelligent systems and network intelligence, robot control, and intelligent optimization of textile industrial process.

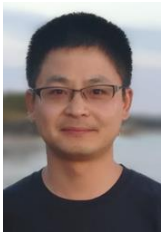

**Dr. Lei Gao** is a senior research scientist in Australia's Commonwealth Scientific and Industrial Research Organization (CSIRO) and a PhD supervisor at both University of Western Australia and Deakin University. Dr. Gao was awarded his BS and PhD degrees in Electrical Engineering from Donghua University, China, in 2001 and 2006, respectively. He published over 100 papers, mostly in prestigious journals and conferences, such as Nature (as the co-lead author and corresponding author). He was awarded the Early Career Research Excellence Prize of the Modelling and Simulation Society of Australia and New Zealand, CSIRO Chairman's Medal for Science Excellence, and Julius Career Award (for exceptional early to mid-career scientists in CSIRO). His research interests include complex system modelling and optimisation, big data modelling, machine learning, robust decision-making, and computational sustainability.

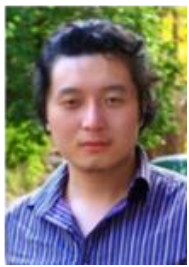

**Xue-song Tang** is currently a Lecturer at College of Information Sciences and Technology, Donghua University, Shanghai, China. He obtained his Ph.D. degree in Computer Science from Fudan University, Shanghai, China in 2015. His scientific interests include deep learning, machine vision, image processing, inter-discipline research on brain science and artificial intelligence.

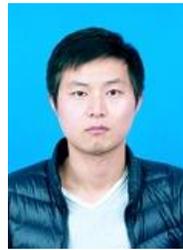

**Bing Wei** is currently a Lecturer at College of Information Sciences and Technology, Donghua University, Shanghai, China. He obtained his Ph.D. degree in College of Information Sciences and Technology, Donghua University, Shanghai, China. He has been a Joint Ph.D. Student with the Department of Electrical, Computer, and Biomedical Engineering, University of Rhode Island, Kingston, RI, USA, from 2018 to 2019. His scientific interests include biological computing, deep learning, image processing and artificial intelligence